\documentclass[10pt,twocolumn,letterpaper]{article}

\usepackage{verbatim}
\usepackage{iccv}
\usepackage{times}
\usepackage{epsfig}
\usepackage{graphicx}
\usepackage{amsmath}
\usepackage{amssymb}
\usepackage{adjustbox}
\usepackage{colortbl}

\usepackage{pifont}
\usepackage[square, sort, comma, numbers]{natbib}
\makeatletter
\@namedef{ver@everyshi.sty}{}
\makeatother
\usepackage{algorithm2e}
\usepackage[breaklinks=true,bookmarks=true,colorlinks=true,linkcolor=blue]{hyperref}
\usepackage{tikz}
\usepackage{pgfplots}
\pgfplotsset{compat=1.15}
\usepackage{amssymb}
\usepackage{calc}
\usetikzlibrary{positioning, calc, arrows, arrows.meta}
\usepgfplotslibrary{groupplots}

\tikzstyle{textbox}=[align=center, outer sep=0pt, minimum size=5pt]
\tikzstyle{fancybox}=[rounded corners, line width=0.8pt]
\tikzstyle{fancyarrows}=[rounded corners, thick, color=black]
\tikzstyle{ff}=[fill=cyan!40!white, fill opacity=1.0, text opacity=1.0]
\tikzstyle{norm}=[fill=yellow!40!white, fill opacity=1.0, text opacity=1.0]
\tikzstyle{att}=[fill=orange!50!white, fill opacity=1.0, text opacity=1.0]
\tikzstyle{inp}=[fill=pink!40!white, fill opacity=1.0, text opacity=1.0]
\tikzstyle{linear}=[fill=blue!30!white, fill opacity=1.0, text opacity=1.0]
\tikzstyle{sigmoid}=[fill=green!30!white, fill opacity=1.0, text opacity=1.0]
\tikzstyle{treenode}=[inner sep=0pt, outer sep=0pt, opacity=0.6]
\tikzstyle{task}=[line width=0.5pt, rounded corners, text opacity=1.0]

\tikzstyle{pred}=[line width=0.5pt, text opacity=1.0]
\tikzstyle{ctrl}=[line width=0.5pt, text opacity=1.0]
\tikzstyle{test} = [draw, circle]
\tikzstyle{task1}=[fill=black!15!blue, fill opacity=0.4, color=black!30!blue]
\tikzstyle{task2}=[fill=black!20!green, fill opacity=0.4, color=black!30!green]
\tikzstyle{task3}=[fill=black!15!red, task, fill opacity=0.4, color=black!30!red]

\tikzstyle{taskctrl1}=[ctrl, draw=black!20!blue, fill opacity=1.0, fill=blue!30!white]
\tikzstyle{taskctrl2}=[ctrl, draw=black!20!green, fill opacity=1.0, fill=green!30!white]
\tikzstyle{taskctrl3}=[ctrl, draw=black!20!red, fill opacity=1.0, fill=red!30!white]

\tikzstyle{taskpred1}=[pred, draw=black!20!blue, fill opacity=1.0, fill=blue!30!white]
\tikzstyle{taskpred2}=[pred, draw=black!20!green, fill opacity=1.0, fill=green!30!white]
\tikzstyle{taskpred3}=[pred, draw=black!20!red, fill opacity=1.0, fill=red!30!white]

\tikzstyle{positive}=[fill=black!15!green, fill opacity=0.5]
\tikzstyle{negative}=[fill=black!15!red, fill opacity=0.5]
\tikzstyle{neutral}=[fill=lightgray]
\tikzstyle{softbox}=[line width=0.8pt, dashed, rounded corners, text opacity=1.0, color=lightgray]
\tikzstyle{softarrow}=[line width=0.8pt, dashed, color=lightgray]

\tikzstyle{featbox}=[rectangle, minimum width=1em, minimum height=1em]
\tikzset{%
  do path picture/.style={%
    path picture={%
      \pgfpointdiff{\pgfpointanchor{path picture bounding box}{south west}}%
        {\pgfpointanchor{path picture bounding box}{north east}}%
      \pgfgetlastxy\x\y%
      \tikzset{x=\x/2,y=\y/2}%
      #1
    }
  },
  sin wave/.style={do path picture={    
    \draw [line cap=round] (-3/4,0)
      sin (-3/8,1/2) cos (0,0) sin (3/8,-1/2) cos (3/4,0);
  }},
  cross/.style={do path picture={    
    \draw [line cap=round] (-1,-1) -- (1,1) (-1,1) -- (1,-1);
  }},
  plus/.style={do path picture={    
    \draw [line cap=round] (-3/4,0) -- (3/4,0) (0,-3/4) -- (0,3/4);
  }}
}

\def\vertspace{15pt}
\def\horspace{10pt}

\iccvfinalcopy 

\def\iccvPaperID{10039} 

\ificcvfinal\fi

\newcommand\tick{\checkmark}
\newcommand\untick{\ding{55}}

\def\*#1{\mathbf{#1}}

\newcommand\mlp{\text{MLP}}
\newcommand\mha{\text{MHA}}
\newcommand\lnorm{\text{LN}}
\newcommand\sigmoid{S}
\newcommand\sa{\text{SA}}
\newcommand\ca{\text{CA}}
\newcommand\Q{\*Q}
\newcommand\K{\*K}
\newcommand\V{\*V}

\newcommand\shcite{(missing quote)}

\begin{document}

\title{Multi-label classification with Transformers for Action Unit Detection}

\author{Gauthier Tallec\\
ISIR\\
\and
Edouard Yvinec\\
ISIR/Datakalab\\
\and
Arnaud Dapogny\\
Datakalab\\
\and
Kevin Bailly\\
ISIR/Datakalab\\
\\
}

\maketitle
\ificcvfinal\thispagestyle{empty}\fi


\begin{abstract}
Action Unit (AU) Detection is the branch of affective computing that aims at recognizing unitary facial muscular movements. It is key to unlock unbiased computational face representations and has therefore aroused great interest in the past few years. One of the main obstacles toward building efficient deep learning based AU detection system is the lack of wide facial image databases annotated by AU experts. In that extent the ABAW challenge paves the way toward better AU detection as it involves a ~2M frames AU annotated dataset. In this paper, we present our submission to the ABAW3 challenge. In a nutshell, we applied a multi-label detection transformer that leverage multi-head attention to learn which part of the face image is the most relevant to predict each AU. 
\end{abstract}

\begin{figure}
\centering
\input{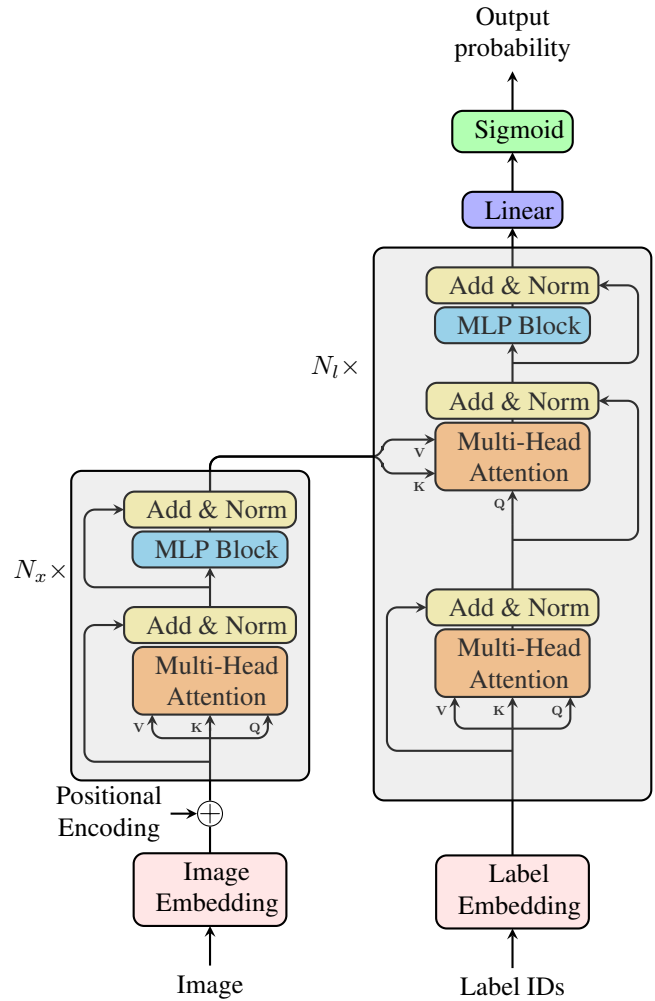}
\caption{Overview of the Multi-Label Transformer \label{fig:overview}}
\end{figure}


\section{Introduction}
The ABAW3 Challenge \cite{kollias2022abaw} is the third edition \cite{kollias2020analysing, kollias2021analysing, kollias2022abaw} of Affect Analysis in the wild \cite{zafeiriou2017aff}. We participated in the AU detection part of the challenge. In previous editions, top ranked methods took benefit from the heterogeneous annotations of ABAW3 \cite{kollias2021distribution, kollias2021affect, kollias2019expression, kollias2019deep, kollias2019face} and used multi-task learning to jointly learn the different affective computing tasks corresponding to the different challenges. This year, the use of such strategy was restricted to the multi-task challenge competitors. We therefore focused on the AU detection problem specifities. 

Action Units (AUs) are a dictionary of unitary muscular activations that was designed by psychologists to anatomically describe the mechanics of facial expressions. Theoretically \cite{ekman1997face}, accurate detection of such activations enables unbiased computational description of human faces and could therefore help improve face analysis applications altogether.

From a machine learning standpoint, AU detection is a multi-label detection problem that comes with three main specifities: First, AUs activations are very local changes of skin texture. Consequently, several methods tried to design locally adapted features \cite{zhao2016deep} with a notable focus on facial landmark-based AU attention \cite{li2017eac, shao2020jaa}. Second, AUs are heavily dependant from one another (mostly for physiological reasons), hence multiple approaches tried to model those dependencies using label prior dependency learning methods \cite{tallec2022multi, song2021hybrid}. Finally, AU are events of short duration, therefore video-based datasets, which include all publicly available AU detection datasets, are bound to be heavily imbalanced toward the absence of activation. This problem has mostly been tackled by weighting the binary cross-entropy with AU frequency based coefficients \cite{shao2019facial} and adding a dice score contribution to the loss \cite{shao2020jaa}.

In this paper, we design a transformer-based architecture for multi-label detection that we apply to the action unit detection problem. More precisely, our method learns local features adapted to each AU using a) self-attention on encoded patches of human faces to produce the features and b) cross-attention between those features and learned AU token for selecting which feature are most informative about each AU. Figure \ref{fig:overview} summarizes the architecture of our model. 
\label{sec:introduction}


\section{Related Work}
\subsection{Action Unit Detection.}
The research work in action unit detection is centered around two main streams, that corresponds to the two main specificities of the problem, namely the local aspect of AUs and the strong dependencies that AUs displays between one another.

Exploiting the local aspect of action units require the extraction of local features. For that purpose, seminal work \cite{zhao2016deep} proposed convolutional layers with region-wise shared filters to extract features that are adapted to each face region. Nevertheless, for such method to be efficient face parts need to always fall in the same rectangular region. To overcome this limitation, several works \cite{li2017eac, shao2018deep, shao2020jaa} guided AU related feature using facial landmarks, either by learning from predefined landmark neighborhood crops \cite{li2017eac}, or by using landmark based AU attention maps \cite{shao2019facial}.

To take advantage of the strong relationships between AU, several works focused on explicitly modeling those dependencies. In particular, backpropagation through a probabilistic graphical model (PGM) was adopted in \cite{Corneanu_2018_ECCV}, and an hybrid message passing strategy was used in \cite{song2021hybrid}. Others leveraged the local nature of AU to assume that label dependencies imply dependencies between local face zones. Attempts at capturing such spatial dependencies include attention map learning \cite{shao2019facial}, LSTM-based spatial pooling \cite{niu2019local} and more recently transformer-like architectures \cite{jacob2021facial}.

\subsection{Transformers for Multi-label Detection.}

The arrival of transformers \cite{vaswani2017attention} has significantly modified the landscape of state of the art deep learning architectures, first in natural language processing \cite{devlin2018bert} and more recently in computer vision \cite{dosovitskiy2020image}. So far, transformer-based multi-label leveraged cross-attention between label-wise learnable tokens \cite{liu2021query2label} and encoded input parts to learn which portions of the inputs are useful for predicting each label. In this work, we propose to apply this very architecture to AU detection. 


\section{Methodology}
\subsection{Background}
Most transformer architectures consists in a combination of Multi-Head Attention modules (MHA) and Multi-Layer Perceptron (MLP) blocks interspersed by layer normalization \shcite and dropout \shcite denoted $\lnorm$ and $\*D$ respectively.

\textbf{The Multi-Head Attention module} consists in an ensemble of $N_h$ attention head. Each of these heads computes attention between $n_q$ queries $\mathbf{q}$ stored in $\Q \in \mathbb{R}^{n_q \times d}$ and $n$ keys $\K \in \mathbb{R}^{n \times d}$ for combining $n$ values $\V \in \mathbb{R}^{n \times d}$. For that purpose, for each head $h$, query, key and values are encoded using dense layers $\*W_{\Q}^{(h)} \in \mathbb{R}^{d \times d_k}, \*W_{\K}^{(h)} \in \mathbb{R}^{d \times d_k}$ and $\*W_{\V}^{(h)} \in \mathbb{R}^{d \times d_v}$:
\begin{equation} 
\tilde{\Q}^{(h)} = \Q \*W_{\Q}^{(h)}, \tilde{\K}^{(h)} = \mathbf{\K} \*W_{\K}^{(h)}, \tilde{\V}^{(h)} = \V \*W_{\V}^{(h)}.
\end{equation}
Then each head performs encoded values combination based on encoded key/query comparison:
\begin{equation}
    \*H^{(h)} = \text{softmax}\left(\frac{\tilde{\Q}^{(h)}(\tilde{\K}^{(h)})^{T}}{\sqrt{d_k}}\right)\tilde{\V}^{(h)}.
\end{equation}

The $N_h$ resulting vectors are concatenated and projected using dense layers $\*W_{O} \in \mathbb{R}^{N_h  d_v \times d}$:
\begin{equation}
    \mha(\Q, \K, \V ; \*M) = \text{Concat}(\*H^{(1)}, \dots, \*H^{(N_h)})\*W_{O}.
\end{equation}
To simplify the notations, we use $d_k = d_v = d  / N_h$ \shcite and further define self and cross attention layers as follows:
\begin{align*}
    \sa(\*Q; \*M) &= \mha(\Q, \Q, \Q)\\
    \ca(\Q; \K; \*M) &= \mha(\Q, \K, \K)
\end{align*}

\textbf{The MLP Block} is composed of two dense layers with dropout on top of each of them: $\*W_{g} \in \mathbb{R}^{d \times d_{mlp}}$ with gelu activation \shcite and $\*W_{l} \in \mathbb{R}^{d_{mlp} \times d}$ with linear activation. Formally for input queries $\*Q$:
\begin{equation}
    \*Q_{mlp} = \*D(\text{GELU}(\*Q \*W_g)),
    \mlp(\*Q) = \*D(\*Q_{mlp} \*W_l).
\end{equation}

\subsection{Multi-Label Transformer}
\subsubsection{Self Attention on input image}
Recent work \cite{dosovitskiy2020image} on transformer architectures unlocked the use of self attention mechanism on images. However it has been shown that applying such mechanism on raw images requires large training datasets with a lot of variability which is only partially the case for ABAW3 database. For that purpose we use a pretrained encoder $g_{\*W_I}$ to first encode the raw images $\*X$ in a patch-based fashion:
\begin{equation}
    \tilde{\*X}^{(0)} = g_{\*W_I}(\tilde{\*X}) \in \mathbb{R}^{n_x \times d},
\end{equation}
Second we encode positions by adding a different token to each encoded patch: 
\begin{equation}
    \*X^{(0)} = \tilde{\*X}^{(0)} + \*I_{n_x}\*W_{n_x}
\end{equation}
where $\*W_{n_x} \in \mathbb{R}^{n_x \times d}$ stores each patch position embedding. Finally, we apply $N_x$ layers of self-attention on the extracted patches. Layer $l$ consists in first, input self attention:
\begin{equation}
    \*X^{(l)}_{xx} = \sa_{xx}^{(l)}(\*X^{(l)}),
    \tilde{\*X}^{(l)}_{xx} = \lnorm[\*X^{(l)} + \*D(\*X^{(l)}_{xx})],
\end{equation}
followed by MLP block based encoding:
\begin{equation}
    \*X^{(l)}_{mlp} = \mlp^{(l)}(\tilde{\*X}^{(l)}_{xx}),
    \*X^{(l + 1)} = \lnorm[\tilde{\*X}^{(l)}_{xx} + \*D(\*X^{(l)}_{mlp})].
\end{equation}

\subsubsection{Cross Attention on label tokens}
Previous works \cite{liu2021query2label} on multi-label transformers leverage $N_l$ layers of multi-head attention between learnable label tokens and input parts in order to learn which portion of the input is the more informative about each label. Formally, we denote $\*T^{(0)} = \*I_{L} \*W_t \in \mathbb{R}^{L \times d}$ the $L$ labels ids encoded with dense layer $\*W_t \in \mathbb{R}^{L \times d}$ and $\*X^{(N_x)}$ the $n_x$ considered parts extracted from input. Layer $l$ consists in, first,  token self attention:
\begin{equation}
    \*T^{(l)}_{tt} = \sa_{tt}^{(l)}(\*T^{(l)}),
    \tilde{\*T}^{(l)}_{tt} = \lnorm[\*T^{(l)} + \*D(\*T^{(l)}_{tt})].
\end{equation}
Second, token image cross attention:
\begin{equation}
    \*T^{(l)}_{tx} = \ca_{tx}^{(l)}(\tilde{\*T}_{tt}^{(l)}, \*X^{(N_x)}), \tilde{\*T}^{(l)}_{tx} = \lnorm[\tilde{\*T}^{(l)}_{tt} + \*D(\*T^{(l)}_{tx})],
\end{equation}
and finally, MLP block based encoding:
\begin{equation}
    \*T^{(l)}_{mlp} = \mlp^{(l)}(\tilde{\*T}^{(l)}_{tx}),
    \*T^{(l + 1)} = \lnorm[\tilde{\*T}^{(l)}_{tx} + \*D(\*T^{(l)}_{mlp})].
\end{equation}
At the end of the $N_l$-th layer, the vector of predictions is computed by projecting the $L$ encoded tokens using a dense layer $\*W_p \in \mathbb{R}^{d \times 1}$ with sigmoid activation $\sigma$:
\begin{equation}
\*p = \sigmoid(\*T^{(N_l)} \*W_p) \in \mathbb{R}^{T}.
\label{eq:prediction}
\end{equation}
Task $t$ distribution is then estimated as follows:
\begin{equation}
\begin{aligned}
    \log p(y^{(t)} \mid \*x, \*W) &= - \text{BCE}(y^{(t)}, p^{(t)}), \\
                                             &= y^{(t)} \log p^{(t)} + (1 - y^{(t)}) \log (1 - p^{(t)}),
\end{aligned}
\end{equation}
where BCE stands for binary cross entropy and $\*W$ englobes all the network parameters. 

Training is done under the assumption that tasks are independent given the input. Therefore it consists in the minimization of the following maximum likelihood based loss:
\begin{equation}
    \mathcal{L}(\* W) = \sum_{t=1}^{T} \text{BCE}(y^{(t)}, p^{(t)}).
    \label{eq:bce}
\end{equation}

Lastly, the ABAW3 challenge is based on videos, \textit{i.e.} on continuous frame sequences. As AUs are segment-level events as opposed to frame-level events (\textit{i.e.} activation of a specific AU usually lasts for at least a couple of frames): to impose such structure in the transformer prediction, we simply smoothed these predictions using a moving mean of window size $w_s = 20$ frames.

\label{sec:methodology}


\section{Experiments}
\label{sec:experiment}

\begin{table*}
\centering
    \label{tab:abaw3_submissions}
    \begin{tabular}{|c|c|c|c|c|c|c|c|}
\hline
Submission ID & moving mean & frequency weights & Dice & DISFA/BP4D pretraining & mixup & Valid F1 & Test Mean F1 \\ 
\hline
1 & \untick & \tick & \tick  & \untick &  \tick & 52.7 & - \\
2 & \tick & \tick & \tick  & \untick &  \tick & 53.8 & - \\
3 & \tick & \untick & \untick  & \tick & \untick & 50.3 & - \\
4 & \tick & \untick & \tick  & \tick &  \untick & 51.0 & - \\
5 & \tick & \untick & \tick  & \tick &  \tick & 51.7 & - \\
\hline
\end{tabular}
    \caption{Summary of submissions to the ABAW3 Challenge}
\end{table*}

\subsection{Network Architecture}
For all our experiments we use an (Imagenet-pretrained) InceptionV3 backbone from which the pooling layer is removed and replaced by a 1D convolution with $128$ output channels, on top of which we add a single self-attention layer with $N_h = 8$. As far as cross attention layers are concerned we use $N_l = 2$ as well as a dropout rate of $0.1$.

\subsection{Training Strategy}
Actions units detection is a heavily imbalanced problem, therefore we use a number of solutions to mitigate this problem while learning: we weight each example (term of the sum in Equation \ref{eq:bce}) with frequency based coefficients. Furthermore, we add a Dice score contribution \cite{shao2020jaa} to the final loss in order to adapt to the F1Score based evaluation.

For loss optimization we use AdamW optimizer \cite{adamw}. For the Inception part of the network we use an exponentially decaying learning rate with initial value $5e-4$ and decay rate $0.99$. For the transformer part of our network we use the learning schedule in \cite{vaswani2017attention}. We scale it with respect to the number of patches $n_x=64$ in the input image self attention part, and with respect to the number of labels $L$, in the label tokens cross-attention part.

\subsection{Pretraining and Data Augmentation}
We investigate pretraining the network on already available data, e.g. BP4D \cite{zhang2013high} and DISFA \cite{mavadati2013disfa} databases. However, these datasets do not have exatcly the same AU annotation as in ABAW3 dataset: Hence, we handle this by pretraining on the reunion of the 3 different AU sets and masking the bce loss for the AU that are not annotated for each example of each dataset. 

Furthermore, we apply data-augmentation by combining geometrical (random rotation, horizontal flips, random zoom) and color-based (channel drop, random brightness) augmentations. Furthermore we make use of label smoothing \cite{labelsmoothing} and image mixup \cite{zhang2018mixup}.

Table \ref{tab:abaw3_submissions} reports the results of our different submissions to the challenge.


\section{Acknowledgements}

This work was granted access to the HPC resources of IDRIS under the allocation 2021-AD011013183 made by GENCI.
\label{sec:conclusion}

{\small
\bibliographystyle{ieee_fullname}
\bibliography{main}
}

\end{document}